\documentclass[letterpaper, 10 pt, conference]{ieeeconf}
\IEEEoverridecommandlockouts                               
\usepackage[hidelinks,bookmarks=false]{hyperref}  % load AFTER pdfx

\usepackage{breqn}
\usepackage{cuted}
\usepackage{xcolor}
\usepackage{capt-of}
\usepackage{amsfonts}
\usepackage{hyperref}
\usepackage{multirow}
\usepackage{mathtools}
\usepackage{dblfloatfix}
\usepackage{amssymb}
\usepackage{ulem}
\usepackage{amsmath}
\usepackage{bbm}
\usepackage{float}
\usepackage{algorithmic}
\usepackage{booktabs}  
\usepackage{tabularx}  
\usepackage{gensymb}  
\usepackage{hyperref}  
\usepackage{cleveref}  
\usepackage[font=small, labelfont=bf]{caption}
\usepackage[ruled, lined, linesnumbered, commentsnumbered, longend, noend]{algorithm2e}
\usepackage{subcaption}
\usepackage{dblfloatfix} % figure[H]
\usepackage{graphicx}
\usepackage{pifont}
\usepackage{mycomments}
\usepackage[size=footnotesize]{caption}

\let\oldding\ding %Store old \ding in \oldding
\renewcommand{\ding}[2][1]{\scalebox{#1}{\oldding{#2}}} %Scale \oldding via optional argument
\definecolor{pink}{RGB}{255, 192, 203}

\def\commenton{1}  % Comment out this line to hide comments
\def\editon{1}     % Comment out this line to hide deletions and coloring of additions{}

\def \paperName{HP$^2$-SLAM}

\title{\LARGE \bf \paperName: Adaptive Hybrid ICP for\\ Robust and Efficient LiDAR SLAM}

\author{%
    Nam Tran$^{1,\dagger}$,
    Thu Tran$^{1,\dagger}$,
    Hieu Phan$^{1,\dagger}$,
    Thai Luu$^{1,\dagger}$,
    Toan Nguyen$^{2}$,
    William~J.~Beksi$^{3}$,
    and~Tuan~Dang$^{1,*}$%
    \thanks{$^{1}$Department of Electrical Engineering and Computer Science, University of Arkansas, Fayetteville, AR, USA; $^{2}$VinRobotics; $^{3}$Department of Computer Science, University of Texas at Dallas, Richardson, TX, USA. $^{\dagger}$Contributed to this work as summer interns at the Cognitive Robotics Lab, University of Arkansas. $^{*}$Corresponding author: Tuan Dang (\texttt{tuand@uark.edu}).}%
}

\begin{document}

\maketitle 

\thispagestyle{empty}
\pagestyle{empty}

\begin{abstract}
Achieving robustness, accuracy, and efficiency simultaneously remains a central
challenge in light detection and ranging (LiDAR) simultaneous localization and
mapping (SLAM). While learning-based approaches deliver strong benchmark
performance, they often require extensive training, substantial computational
resources, and struggle to generalize to unseen or degenerate environments.
Geometry-based methods are efficient and interpretable, yet their performance
degrades in planar or repetitive scenes due to limitations of standard iterative
closest point (ICP) formulations. We present \paperName~, a minimalist yet
robust LiDAR SLAM framework built around a neighborhood-size adaptive hybrid
ICP. Our key insight is a planarity-aware adaptive threshold that dynamically
classifies correspondences based on local geometric structure and density,
thereby enabling a principled balance between point-to-plane and point-to-point
residuals. This formulation stabilizes alignment in both structured and
degenerate environments without feature engineering, learning modules, or
dataset-specific tuning. Integrated into a complete SLAM pipeline with submap
management, loop closure detection, and pose graph optimization,
\paperName~consistently outperforms strong geometry-based baselines across
publicly available datasets while maintaining real-time performance on commodity
hardware. Our results demonstrate that carefully designed geometric adaptation
can achieve strong generalization and robustness without sacrificing simplicity
or efficiency.
\end{abstract}

% \cm {\textbf{Thu} upload video here, please check it out and give comments: \href{https://drive.google.com/file/d/1MaNX-SX_CWtpMqOpVSsvBC4BG2RahzlA/view?usp=sharing}{https://drive.google.com/file/d/1MaNX-SX_CWtpMqOpVSsvBC4BG2RahzlA/view?usp=sharing}} \\

% \cm{Check out the latest video from \textbf{Thu} at (12:20 PM CST) : \href{https://drive.google.com/file/d/1Yx8pDdeW58sHXp1Anh7diP3NN_8lwe79/view?usp=sharing}{ClickHere}\\
% Video comments:\\
% Looks good!}

\section{Introduction}
\label{sec:introduction}
Simultaneous localization and mapping (SLAM) is a fundamental capability for
autonomous robots operating in unknown environments. Learning-based SLAM systems
such as DeepFactors~\cite{czarnowski2020deepfactors},
DROID-SLAM~\cite{teed2021droid}, and NICE-SLAM~\cite{zhu2022nice} have recently
demonstrated strong benchmark performance. However, these methods often require
extensive training data, high computational resources, and careful domain
adaptation. Their generalization to unseen environments remains limited, and
their black-box nature complicates deployment on resource-constrained robotic
platforms.

Geometry-based light detection and ranging (LiDAR) SLAM remains attractive due
to its efficiency, interpretability, and cross-domain robustness. Classical
systems such as LOAM~\cite{zhang2014loam}, A-LOAM~\cite{aloam2018}, and
LeGO-LOAM~\cite{shan2018lego} rely on feature extraction and scan-to-map
registration. More recently, minimalist approaches such as
KISS-ICP~\cite{vizzo2023kiss} and KISS-SLAM~\cite{guadagnino2025kiss} show that
competitive performance can be achieved without handcrafted features or learning
modules. Nevertheless, a fundamental limitation persists: ICP degeneracy in
planar or repetitive environments. On highways, bridges, and long corridors, 
point-to-point ICP~\cite{li2020evaluation} becomes ill-conditioned, causing drift.
Point-to-plane ICP~\cite{low2004linear} improves convergence in structured
regions, but is sensitive to noise and unreliable when geometric support is
weak.

\begin{figure}[t]
\centering
\includegraphics[width=1.0\columnwidth]{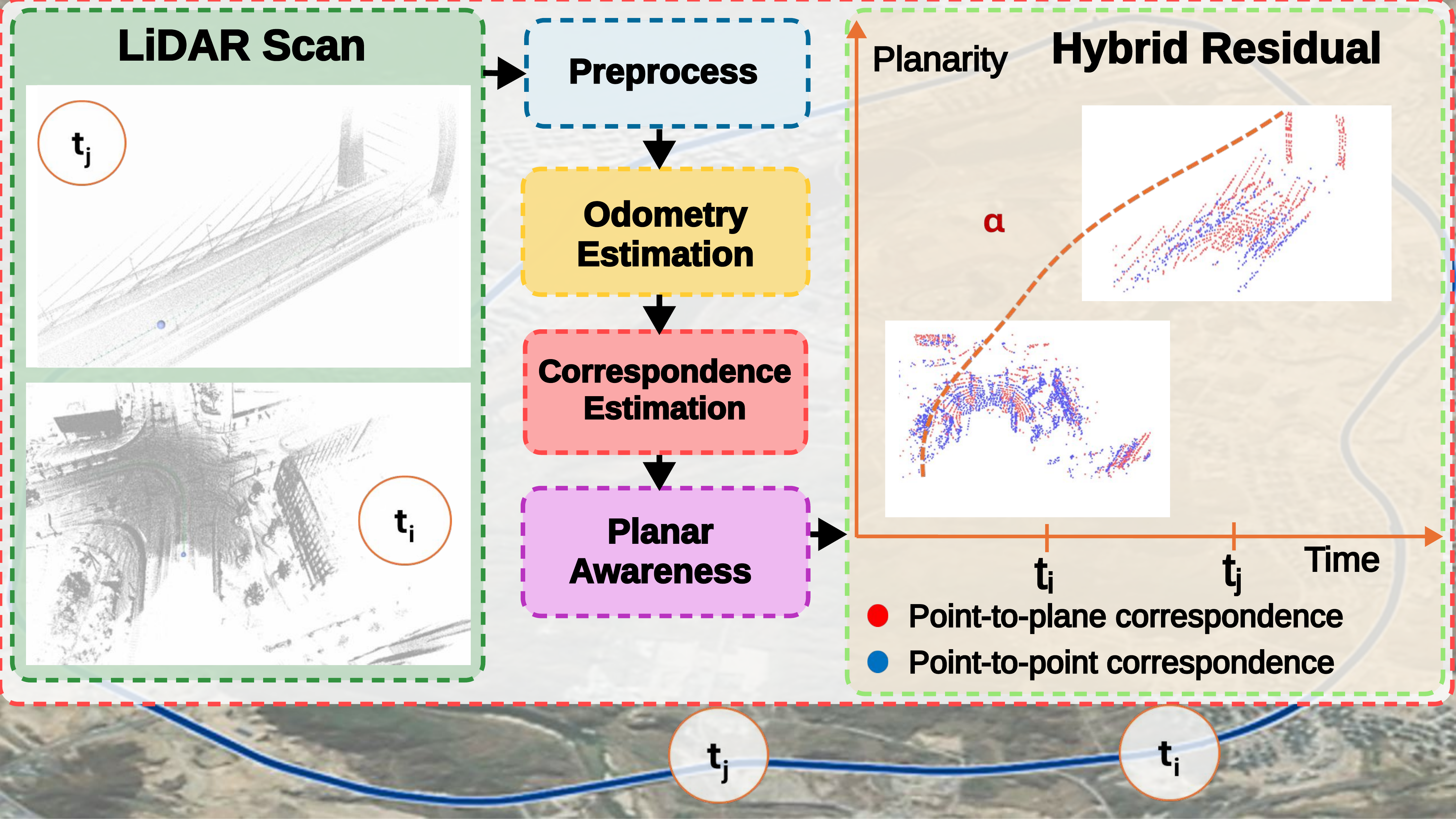}
\caption{\paperName~conceptualization. Each LiDAR scan is registered to a local
map using a density-aware planarity criterion that separates planar and
non-planar correspondences. The hybrid ICP weight $\alpha$ is adaptively updated
at each scan based on the observed planar ratio, increasing in structured
environments and decreasing in less planar settings.
% \del{ This dynamic balancing stabilizes the hybrid objective across diverse
% scenarios.}
}
\label{fig:concept}
\end{figure}

Various hybrid strategies have attempted to address the issue of ICP degeneracy 
but have introduced other drawbacks. Methods such as MULLS~\cite{pan2021mulls} 
and GICP~\cite{segal2009generalized} combine residual formulations but often 
require complex modeling or sensitive parameter tuning. GenZ 
ICP~\cite{lee2024genz} introduces adaptive residual weighting based on 
degeneracy metrics and achieves strong odometry performance. However, its 
adaptation is driven by global degeneracy indicators rather than explicit 
correspondence-level geometric classification, and it focuses on front-end 
odometry without integration into a complete SLAM framework.

This motivates a central question: \textit{Can residual balancing be derived
directly from local geometric structure in a density-aware and principled
manner, while preserving minimalist design?} To this end, we propose
\paperName, a lightweight LiDAR SLAM framework built around a neighborhood-size
adaptive hybrid ICP. As shown in Fig.~\ref{fig:concept}, instead of fixed
thresholds or heuristic weighting, we introduce a density-aware adaptive
planarity criterion that scales with local neighborhood size. Correspondences
are classified into planar and non-planar sets, and the balance between
point-to-plane and point-to-point residuals emerges directly from real-time
geometric observations. This reduces parameter sensitivity and improves
stability in both structured and degenerate environments.

Building on KISS-SLAM~\cite{guadagnino2025kiss}, we integrate this adaptive
hybrid ICP into a complete pipeline that includes submap management, loop
closure detection, and pose graph optimization. This pipeline helps
\paperName~achieve not only accurate odometry but global consistency as well.
Our main contributions are as follows.
\begin{itemize}
  \item A \textbf{neighborhood-size–adaptive planarity criterion} for density-aware correspondence classification.
  \item An \textbf{adaptive hybrid ICP formulation} that dynamically balances residuals based on observed geometry.
  \item A \textbf{minimalist full-SLAM integration} that achieves improved global consistency while preserving real-time efficiency.
\end{itemize}
Extensive experiments on public SLAM datasets show that \paperName~consistently
outperforms strong geometry-based baselines, including
KISS-SLAM~\cite{guadagnino2025kiss}, MULLS~\cite{pan2021mulls},
CT-ICP~\cite{dellenbach2022ct}, and GenZ-ICP~\cite{lee2024genz}, while
maintaining real-time performance and robustness across different types of LiDAR
sensors. We will release our source code to foster reproducibility and
validation, and to contribute to the growth of the research community.

\section{Related Work}
\label{sec:related_work}
% Simultaneous Localization and Mapping (SLAM) has evolved considerably over the last decades.  Bongard, Josh \cite{6792214} points out that SLAM is more difficult than mapping with known poses, since the poses are unknown and must be estimated along the way. Early approaches, such as EKF-SLAM and Graph-SLAM \cite{article}, provided the foundation for probabilistic formulations. Since then, the field has diversified into algorithms tailored to various sensing modalities and application domains. Advances in sensor technology and optimization, together with learning-based methods, have broadened the design space. In what follows, we'd like to categorize related work into LiDAR-based SLAM, with a particular focus on ICP-driven odometry and hybrid alignment strategies.  

\subsection{LiDAR-Based SLAM and ICP-Based Odometry}  
\label{subsec:lidar_based_slam_and_icp-based_odometry} 

LiDAR-based SLAM is widely used for large-scale robotic navigation due to its accurate and illumination-insensitive range sensing. Classical systems such as LOAM~\cite{zhang2014loam} and its variants~\cite{aloam2018,shan2018lego} rely on handcrafted geometric features and careful parameter tuning for scan-to-map registration. More recent minimalist approaches, like KISS-ICP~\cite{vizzo2023kiss} and KISS-SLAM~\cite{guadagnino2025kiss}, demonstrate that competitive performance can be achieved using pure geometric registration without feature extraction or auxiliary sensors. Most of these methods rely on the ICP algorithm, typically adopting either a point-to-point or point-to-plane formulation. The point-to-point formulation is efficient but becomes ill-conditioned in planar or repetitive environments. The point-to-plane formulation \cite{makovetskii2018point,wei2022gclo} improves convergence in structured scenes but demands reliable normal estimation and sufficient geometric support. Recent analyses~\cite{hatleskog2024probabilistic} highlight the complementary strengths and weaknesses of these residuals under degeneracy. The point-to-plane formulation was originally introduced by Chen and Medioni~\cite{chen1992object}, constraining optimization along the normal plane to address slow convergence, while Felix et al.~\cite{felix2025lidar} further refined alignment by minimizing the distance from source points to the tangent planes of corresponding target points within a hybrid-ICP framework. However, existing systems usually employ a single residual globally or combine them using fixed heuristics, limiting their ability to adapt to changing geometric conditions along a trajectory. This limitation motivates our work, in which we introduce a neighborhood-size–adaptive planarity criterion that classifies correspondences based on local geometric structure and density. Defining a planarity criterion enables dynamic balancing between point-to-point and point-to-plane residuals within a unified SLAM framework, thereby improving robustness while preserving computational simplicity.

\subsection{Hybrid Alignment Strategies}  
\label{subsec:hybrid_alignment_strategies}  
% \cm{Make this section shorter and position against GenZ-ICP: (1) odometry only vs Full Slam, (2) Adaptive weighting via degeneracy metrics vs Adaptive classification via geometry, (3) Weighting continuous vs Binary but adaptive} \\
% \cm {\textbf{Nam Tran, Toan Nguyen}: You should check again the above points} \\
To address the limitations of individual error metrics, researchers introduced
hybrid ICP methods that integrate multiple alignment strategies. \add{For
instance,} MULLS~\cite{pan2021mulls} jointly optimizes point-to-point,
point-to-plane, and point-to-line residuals in a least-squares framework, but
requires complex feature extraction and tuning. Generalized-ICP
(GICP)~\cite{segal2009generalized} introduces a probabilistic formulation that
unifies point-to-point and point-to-plane ICP under local planarity assumptions,
improving accuracy over single-metric approaches, and has been further enhanced
in subsequent works
\cite{koukoulis2024unleashing,choi2022fast,koukoulis2025leveraging}, with
widespread applications in large-scale and underground environments
\cite{kim2023adaptive,tuna2023x}. More recently, GenZ-ICP~\cite{lee2024genz}
adapts the weighting between point-to-point and point-to-plane terms based on
local geometry. This adaptive strategy improves robustness in degenerate
environments, such as long corridors, and enhances performance in diverse
real-world settings.  This survey~\cite{lee2024lidar} highlights the growing trend toward hybrid
metrics that balance efficiency and robustness. Unlike prior hybrid formulations
that rely on fixed or heuristic thresholds for planarity classification, our method introduces an adaptive planarity threshold that dynamically adjusts with
local density. This design enables us to separate planar and non-planar
correspondences more reliably, allowing the balance between point-to-plane and
point-to-point residuals to be derived directly from real-time observations. In contrast to MULLS, we avoid explicit feature extraction and complex parameter
tuning while keeping the design minimalist. Unlike GenZ-ICP, which is limited to
odometry, our method integrates \del{the} hybrid ICP with \del{the}\add{an}
adaptive planarity threshold into a complete SLAM framework that combines local
mapping, loop closure, and global
optimization.

\section{System Overview}
\begin{figure*}[t]
\centering
\includegraphics[width=\textwidth]{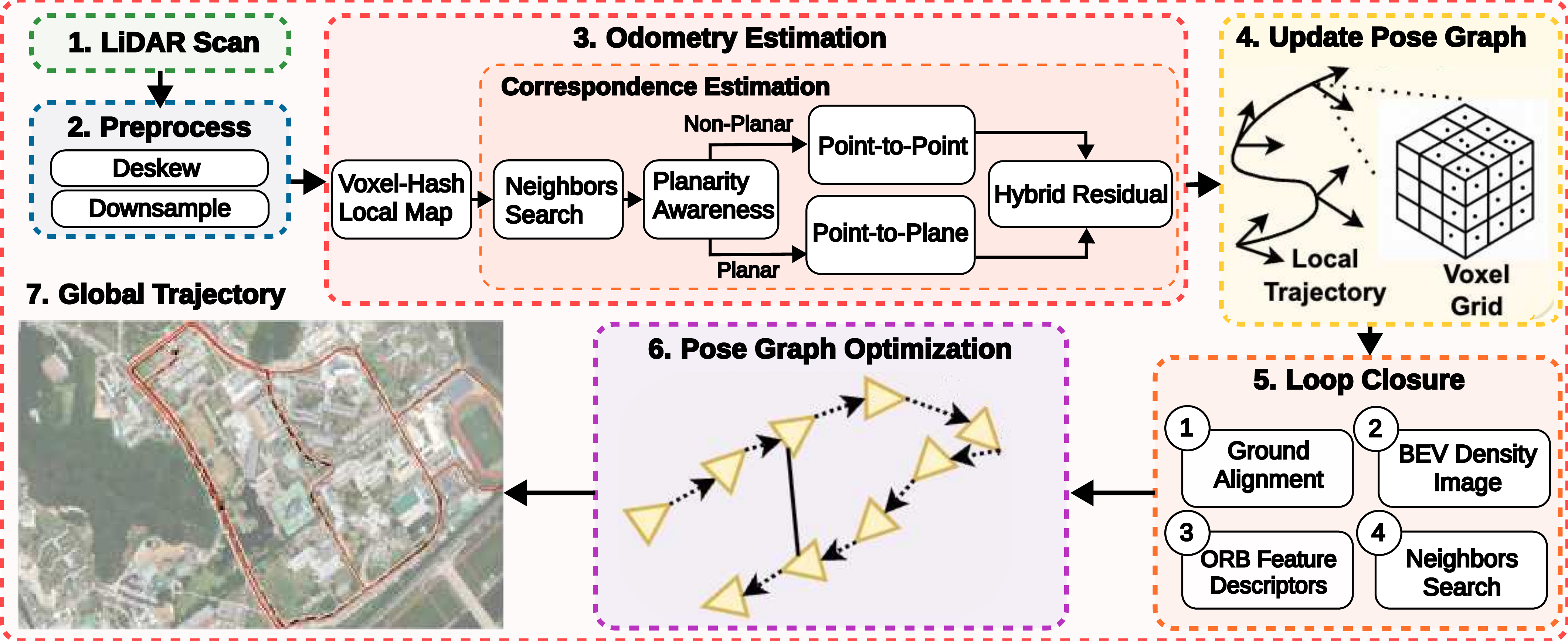}
\caption{An overview of \paperName. Our system includes LiDAR preprocessing,
odometry estimation, local mapping, loop-closure detection using ground-aligned
BEV and ORB with RANSAC verification, and pose-graph optimization for global
consistency. Odometry is improved by a planarity-aware hybrid residual that
applies point-to-plane in planar regions and point-to-point in the non-planar
areas with adaptive weighting. This hybrid approach stabilizes odometry and
reduces longitudinal drift when mapping degenerate environments, such as long
corridors, roads, and bridges.}
\label{fig:overview}
\end{figure*}

We build \paperName~on top of KISS-SLAM, while replacing its
original KISS-ICP module (point-to-point ICP) with a novel \del{A}\add{a}daptive
ICP \add{algorithm}. The objective is to preserve the ``\del{S}\add{s}mall and \del{S}\add{s}imple'' design philosophy of KISS-SLAM, while
improving robustness in geometrically degenerate environments, such as long
corridors or large planar surfaces. \del{The overview of the system is shown in
Figure \ref{fig:overview} and is described as follows:} \add{An overview of our
system is shown in Fig.~\ref{fig:overview} and described below.}

\textbf{LiDAR odometry (adaptive ICP).} Following the original KISS-SLAM
pipeline, each incoming LiDAR scan is first \textit{deskewed} to compensate for
motion distortion and \textit{\del{voxel} downsampled} to reduce point density.
A constant-velocity motion model provides an initial pose guess. The main
difference of our approach lies in the refinement stage. Instead of relying
solely on point-to-point ICP as in KISS-ICP, we introduce an adaptive ICP
mechanism. This mechanism classifies correspondences into two categories: (i)
planar-based methods, optimized using point-to-plane, and (ii) non-planar-based
methods, optimized using point-to-point residuals. An adaptive weight $\alpha$
is determined based on the ratio of planar to non-planar correspondences,
enabling the system to balance the two types of constraints automatically.
After convergence, a downsampled version of the registered scan is integrated
into the current local map.

\textbf{Local mapping.} As in KISS-SLAM, we exploit short-term odometry
consistency by splitting the global map into \textit{submaps}. Each submap $M_k$
is associated with a \textit{keypose} $T_k$, a voxel grid $V_k$, and a local
odometry trajectory. Deskewed and downsampled scans are continuously integrated
into the active submap. Once the traveled distance exceeds a predefined
threshold $\beta$, a new submap is initialized at the current keypose while the
previous submap is finalized and stored in the \textit{pose graph}. In the pose
graph, nodes represent keyposes and edges encode odometry constraints.

\textbf{Loop closure and pose graph optimization.} To reduce accumulated
odometry drift, \del{our system}\add{\paperName} incorporates a loop closure
detection module. Following the method of Gupta et
al.~\cite{gupta2025efficiently}, candidate loop closures are identified by
aligning the current local map point cloud with the $xy$-plane of the keypose
coordinate system. All points are then projected into a bird's-eye-view
\add{(BEV)} representation by computing the density of 3D LiDAR points within
each 2D grid cell. Binary ORB descriptors~\cite{rublee2011orb} are extracted
from these BEV images and matched against previously stored descriptors. To
enable efficient loop-candidate matching, these ORB descriptors are stored and
queried in a Hamming distance embedding binary search
tree~\cite{lim2007hierarchical}. If a potential loop candidate is found,
\add{then} geometric verification is performed using RANSAC\del{,}\add{ to
provide} \del{providing}\del{a}2D alignment between the density images.

To further ensure robustness, we additionally validate loop closures in 3D by
checking the overlap consistency between the registered submaps. Only candidates
with sufficient overlap are kept and added to the pose graph, reducing the risk
of false loop constraints. For global consistency, we optimize the pose graph
using the approach of Grisetti et al.~\cite{grisetti2011tutorial}. In this
graph, keyposes are kept fixed while new nodes and edges are introduced from
odometry and validated loop closures. The optimization is performed at a
fine-grained level, redistributing small odometry errors among scan poses within
local trajectories. This step effectively reduces drift and ensures a globally
consistent map.
\section{Methodology}
\label{sec:methodology}
\subsection{Problem Formulation}
\label{subsec:problem_formuation}
%Bài toán Lidar slam được định nghĩa: Cho chuỗi các point cloud $\{P_t\}^{T}_{t=1}$, trong đó mỗi ${P_t}\in{\mathbb{R}^3}$ là một point cloud quan sát tại thời điểm $t$. Trạng thái của robot tại thời điểm $t$ được biểu diễn bởi 1 ma trận ${T_t}\in{SE(3)}$, với ${T_t} = [R_t|t_t]$, ${R_t}\in{SO(3)}$ và ${t_t}\in{\mathbb{R}^3}$. Mục tiêu bài toán SLAM là đồng thời ước lượng quỹ đạo $X=\{T_t\}^{T}_{t=1}$ và bản đồ môi trường $M\in{\mathbb{R}^3}$. 
The LiDAR SLAM problem is defined as follows\del{:}\add{.} Given a sequence of
3D point clouds $\{P_t\}^{T}_{t=1}$, where each ${P_t}\in{\mathbb{R}^3}$ is a
point cloud observed at time $t$\del{.}\add{,} \del{T}\add{t}he robot state at
\del{time} $t$ is represented by a matrix ${T_t}\in{SE(3)}$, with ${T_t} =
[R_t|t_t]$, ${R_t}\in{SO(3)}$ and ${t_t}\in{\mathbb{R}^3}$. The goal of LiDAR
SLAM is to jointly estimate the trajectory $X=\{T_t\}^{T}_{t=1}$ and a global
map $M$. 

%Trong KISS-SLAM, quá trình căn chỉnh giữa các frame chủ yếu dựa trên point-to-point ICP. Với mỗi cặp điểm tương ứng $(p_i,q_i)$, residual là:
% $e_{pp}(p_i,q_i,R,t)=\|p_i-(Rq_i+t)\|$
%Bài toán tối ưu hóa pose có dạng:
% \[
% \min_{R,\,t} \sum_i \left\| \mathbf{p}_i - \left( R
% \mathbf{q}_i + \mathbf{t} \right) \right\|^2
% \]
%Tuy nhiên như đã thảo luận trong Related Work, point-to-point đơn giản nhưng kém chính xác trong môi trường nhiều mặt phẳng. Ngược lại, point-to-plane khai thác thông tin pháp tuyến:
% $e_{pl}(p_i,q_i,R,t)=n^T_i(p_i-(Rq_i+t))$,
%trong đó n_i là vector pháp tuyến tại p_i.
% Trong bài nghiên cứu của chúng tôi, chúng tôi xây dựng một hàm chi phí kết hợp 2 loại residual thông qua adaptive weight $\alpha$:
% \[
% E(R,t)
% = \alpha \sum_{i \in \mathcal{P}} \rho(e_{\mathrm{pl}}(i)^2)
% + (1-\alpha) \sum_{j \in \mathcal{Q}} \rho(e_{\mathrm{pp}}(j)^2),
% \]
%trong đó:
%- $\mathcal{P}$: là tập điểm planar.
%- $\mathcal{Q}$: là tập điểm non-planar.
%- \alpha\in{[0,1]} được xác định dựa trên tỷ lệ planar trong batch hiện tại.

KISS-SLAM estimates the pose of the current scan by registering it to a
voxelized local map via a point-to-point ICP formulation. \add{Concretely,} \del{G}\add{g}iven a set of
correspondences $\{(p_i,q_i)\}$ between the target and source
point clouds, the point-to-point residual is
\begin{equation}
  e_{{pp}}(p_i,q_i;R,t) = p_i - (Rq_i+t).
\end{equation}
\del{and}\add{The} pose $(R,t)\in SE(3)$ is obtained by minimizing the
sum of squared residuals
\begin{equation}
  \min_{R,t} \ \sum_i \big\|\,p_i - (Rq_i+t)\,\big\|^{2}.
\end{equation}

As \del{reviewed}\add{discussed} in Sec.~\ref{sec:related_work}, the
point-to-point objective is computationally attractive, but can be
ill-conditioned in predominantly planar scenes. A complementary formulation is
the point-to-plane residual\del{, which}\add{that} leverages local surface
normals to enforce orthogonal consistency, i.e.,
\begin{equation}
  e_{pl}(p_i,q_i,R,t)=n^\top_i(p_i-(Rq_i+t)),
  \label{eq:point-to-plan_residual}
\end{equation}
where $n_i$ is the normal vector at $p_i$. 
%\com{Should $\mathbf{n}_i$, $\mathbf{p}_i$, $\mathbf{q}_i$, and $t$ be bold font in \eqref{eq:point-to-plan_residual}?}
In this work, we
define a hybrid alignment objective as a convex combination of point-to-point
and point-to-plane residuals with adaptive weight $\alpha$. To achieve this, we
first analyze the local geometric structure of each correspondence and use it to
balance the two residual terms. For each candidate correspondence, we form a
local map neighborhood $\mathcal{N}$ of size $n=\lvert\mathcal{N}\rvert$ around
the target point and compute the covariance matrix $C$. \add{More formally,}
\del{L}\add{l}et the eigenvalues of $C$ be ordered as $\lambda_1 \ge \lambda_2
\ge \lambda_3$. We define a local planarity score,
\begin{equation}
  s \;=\; \frac{\lambda_3}{\lambda_1+\lambda_2+\lambda_3},
\label{eq:planarity_score}
\end{equation}
such that smaller values indicate a more planar structure.

Rather than employing a fixed cutoff, we define the following
neighborhood-size–adaptive decision boundary that scales inversely with the
local neighborhood size:
\begin{equation}
  \tau(n) \;=\; \tau_{\mathrm{ref}} \,\frac{N_{\mathrm{ref}}}{n}, \qquad
  \tau_{\min} \le \tau(n) \le \tau_{\max}.
\label{eq:adaptive_threshold}
\end{equation}
A correspondence is classified as planar if the local planarity score $s$ falls
below $\tau(n)$, and non-planar otherwise. The evaluation is performed only when
the neighborhood contains at least $n_{\min}$ points, ensuring that the
covariance estimate is statistically meaningful. By construction, $\tau(n)$
decreases with increasing density, enforcing stricter conditions in well-sampled
regions\del{, while}\add{. Conversely,} in sparse neighborhoods it relaxes,
allowing structural cues to be retained even under limited observations. We
denote the resulting planar and non-planar sets as $\mathcal{P}$ and
$\mathcal{Q}$, respectively. 

To ensure practicality, we introduce a reference threshold
$\tau_{\mathrm{ref}}$, which specifies the nominal decision boundary at a
neighborhood size $N_{\mathrm{ref}}$. In practice, $\tau_{\mathrm{ref}}$ acts
only as a coarse scaling factor and can be kept within a narrow range from 0.01
to 0.2 across datasets. Although lowering $\tau_{\mathrm{ref}}$ may yield
further improvements in certain cases, the adaptive formulation remains far less
sensitive to this choice than fixed-threshold schemes. This property
substantially reduces the need for dataset-specific retuning and strengthens the
robustness of our pipeline across diverse sensors and environments.

We formulate the hybrid alignment objective as
% \begin{equation}
%     E(R,t)
% = \alpha \sum_{p \in \mathcal{P}} \rho(e_{\mathrm{pl}}(p)^2)
% + (1-\alpha) \sum_{q \in \mathcal{Q}} \rho(e_{\mathrm{pp}}(q)^2)
% \end{equation}
\begin{align}
  \underset{R,t}{\min} & [\alpha \sum_{(p_i,q_i) \in \mathcal{P}} \rho(e_{\mathrm{pl}}(p_i,q_i, R, t) + \nonumber \\
                       & (1-\alpha) \sum_{(p_i,q_i)\in\mathcal{Q}} \rho(e_{\mathrm{pp}}(p_i,q_i, R, t) ],
\label{eq:error}
\end{align}
where \del{the term} $\rho(\cdot)$ is a robust loss function. 
%\com{Should $\mathbf{n}_i$, $\mathbf{p}_i$, $\mathbf{q}_i$, and $t$ be bold font in \eqref{eq:error}?}
% \cm{Add final optimiation function for $E(R,t$}
The mixing weight $\alpha\in[0,1]$ is set adaptively based on the current planar
ratio, i.e.,
\begin{equation}
  \alpha = \frac{|\mathcal{P}|}{|\mathcal{P}|+|\mathcal{Q}|}.
\end{equation}
In our work, we use the Geman-McClure estimator to suppress the influence of
outliers and dynamic structures. \add{Concretely,} 
\begin{equation}
  \rho(e) = \frac{e^2}{e^2+c^2},
\end{equation}
where $e$ is the residual and $c$ is a scale parameter that controls the degree
of down-weighting for significant errors. 

Unlike the standard least-squares loss, which penalizes large residuals
quadratically and can be overly sensitive to outliers, the Geman-McClure
estimator saturates as $|e|$ grows large. Specifically, when $e$ is small,
$\rho(e) \approx e^2/c^2$ behaves similarly to a squared error, ensuring good
convergence for inlier correspondences. However, for large $|e|$, $\rho(e)$
approaches $1$, effectively down-weighting the contribution of these residuals
in the optimization. This property makes the loss function robust by reducing
the influence of outliers or dynamic structures that would otherwise dominate
the objective and skew the estimated pose. 

In order to avoid manual tuning, we set $c=\sigma$, where $\sigma$ is estimated
online from recent motion prediction errors between the constant-velocity prior
and the registered pose. \add{More formally,}
\begin{equation}
  \mathrm{err}_k = \|\delta t_k\| + 2 r_{\max} \sin(\theta_k/2),
  \label{eq:motion_error}
\end{equation}
where $\delta t_k$ is the translational error, $r_{\max}$ denotes the maximum
LiDAR sensing range, and $\theta_k$ is the rotational error angle. The term $2
r_{\max} \sin(\theta_k/2)$ converts the rotational deviation into an equivalent
displacement at the farthest observed point, ensuring consistent units with the
ICP residuals. The robust scale is then computed as the \del{root-mean-square
(RMS)} \add{root mean square} of the accumulated motion errors over a sliding
window of size $N$,
\begin{equation}
  \sigma = \sqrt{\frac{1}{N} \sum_{k=1}^{N} \mathrm{err}_k^2}.
  \label{eq:sigma_rms}
\end{equation}
This motion-adaptive scale enlarges the inlier region under aggressive motion
and tightens it during slow motion, maintaining consistent robustness across
varying motion regimes. 

% Để phân loại planar và non-plannar, với mỗi tương ứng, xét lân cận $\mathcal{N}$ (kích thước $N=|\mathcal{N}|$) từ bản đồ và tính ma trận hiệp phương sai $C$ để thu được các trị riêng  $\lambda_1\!\ge\!\lambda_2\!\ge\!\lambda_3\!\ge\!0$.
%Độ phẳng cục bộ được đo bởi:
% \[
% s \;=\; \frac{\lambda_3}{\lambda_1+\lambda_2+\lambda_3}.
% \]
%Khác với ngưỡng cố định, chúng tôi sử dụng ngưỡng \emph{thích ứng} theo kích thước lân cận:

% \[
% \tau(n)
% = \tau_{\mathrm{ref}}\;\frac{N_{\mathrm{ref}}}{n},
% \qquad
% \tau_{\min}\le \tau(n)\le \tau_{\max}.
% \]
% Vậy ác tương ứng được gán là \emph{planar} nếu lân cận đủ lớn và độ phẳng $s$ nhỏ hơn ngưỡng thích ứng %\tau(n)%; ngược lại là \emph{non-planar}. $\tau(N)$ \emph{giảm} khi lân cận dày (đòi hỏi bề mặt thật sự phẳng) và \emph{tăng} khi lân cận thưa (khoan dung hơn trước nhiễu/sampling).

\subsection{Implementation Details}
\label{subsec:implementation_details}
%Hệ thống của chúng tôi được xây dựng trên nền tảng KISS-ICP và mở rọng thành KISS-SLAM với cơ chế hybrid aglignment và ngưỡng phân loại mặt phẳng thích nghi. Mỗi khung quét Lidar đầu vào trước tiên được deskew bằng constant-velocity motion model để bù trừ biến dạng cho chuyển động cảm biến.
% \[
%  P_t^{deskew}=Exp((s_i-1){\omega})P_t
% \]
%{s_i}\in[0,1] là thời gian chuẩn hóa của điểm P_t trong frame, $\omega=log(\Delta{T})\in{se(3)}$, \Delta{T} là chuyển động của cảm biến từ đầu đến cuối frame.
%Sau đó dữ liệu được voxelize và downsample để tạo bản đồ cục bộ theo cấu trúc voxel hash structure giúp hỗ trợ tìm kiếm nhanh neighbor cho ICP registration và cập nhật bản đồ. 
\del{Our system}\add{\paperName} is built on KISS-ICP and \del{is} extended into
a KISS-SLAM pipeline with a hybrid alignment mechanism and adaptive planar
classification. Each incoming LiDAR scan is first deskewed using a
constant-velocity motion model to compensate for sensor motion,
\begin{equation}
  P_t^{\mathrm{deskew}} = e^{(s_i-1){\omega}P_t},
\end{equation}
where ${s_i}\in{[0,1]}$ is the normalized timestamp within the scan and
$\omega=log(\Delta{T})\in \mathfrak{se}(3)$, $\Delta{T}\in{SE(3)}$ denotes the
relative transform from start to end in the scan. After deskewing, the scan is
voxelized and downsampled. We maintain a voxel-hashed local map to support fast
nearest-neighbor queries for ICP registration and efficient map updates.
%Với mỗi cặp tương ứng ứng viên, chúng tôi trích xuất một lân cận từ bản đồ voxel và tính toán ma trận hiệp phương sai bằng PCA. Tỉ lệ giá trị riêng cung cấp chỉ số planarity score  theo (6), sau đó được so sánh với một ngưỡng thích nghi (7) phụ thuộc vào kích thước lân cận. Ngưỡng này tự động giảm trong vùng dày đặc và tăng trong vùng thưa thớt. Nhờ vậy, các tương ứng được gán nhãn thành hai loại: planar hoặc non-planar. 

For each candidate correspondence, we retrieve a local neighborhood from the
voxel map and compute the covariance matrix via principal component analysis.
From its eigenvalues, we compute the planarity score $s$ as in Equation
\eqref{eq:planarity_score} and compare it with the adaptive threshold $\tau(n)$
in Equation \eqref{eq:adaptive_threshold}. This adaptive decision boundary increases
tolerance in sparse regions while enforcing stricter flatness conditions in
dense areas. As a result, \add{the} correspondences are automatically categorized into
planar or non-planar classes. 

%Các tương ứng phẳng được sử dụng trong ràng buộc point-to-plane, trong khi các tương ứng phi phẳng sử dụng ràng buộc point-to-point. Hai nhóm này được xử lý song song, đồng thời sử dụng robust kernel để giảm ảnh hưởng của outlier. Hàm mục tiêu lai được cực tiểu hóa bằng phương pháp Gauss–Newton trên đại số Lie se(3), với pose được cập nhật lặp cho đến khi hội tụ. Hệ số cân bằng giữa hai dạng residual được tính động dựa trên tỉ lệ planar/phi phẳng hiện tại, nhằm duy trì sự hài hòa giữa ràng buộc hình học mạnh mẽ và tính ổn định.

Planar correspondences are aligned with a point-to-plane residual, whereas
non-planar correspondences rely on point-to-point distances. Both sets are
processed in parallel, and robust kernels are employed to mitigate the influence
of outliers. The hybrid alignment objective is minimized in a Gauss-Newton
framework \add{based} on the Lie algebra $\mathfrak{se}(3)$, updating the pose
incrementally until convergence. The weighting factor between planar and
non-planar residuals is computed adaptively from the current ratio of classified
correspondences, balancing geometric constraints with robustness.

%Sau khi căn chỉnh, bản đồ cục bộ được cập nhật với khung quét mới đã đăng ký. Để bảo đảm tính nhất quán toàn cục, chúng tôi tích hợp thêm mô-đun loop closure dựa trên chiếu BEV (bird’s eye view). Mỗi bản đồ cục bộ được chiếu xuống mặt phẳng mặt đất, từ đó trích xuất đặc trưng ORB và lưu vào cơ sở dữ liệu. Các ứng viên vòng lặp được xác minh bằng căn chỉnh 2D RANSAC trước khi được thêm vào đồ thị pose toàn cục. Tại đây, tối ưu hóa pose graph giúp hiệu chỉnh drift theo thời gian và bảo toàn tính nhất quán bản đồ khi quay lại các khu vực đã thăm.
\begin{figure}[!t] %Fig3
\centering
\includegraphics[width=1\linewidth]{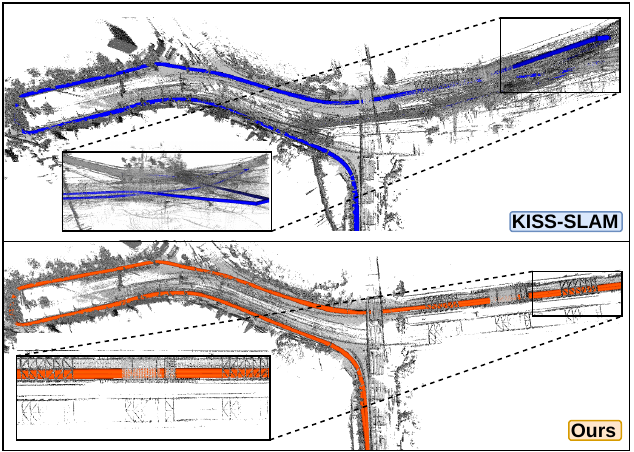}
\caption{Mapping results on the HeLiPR Bridge (Avia04) sequence. Our method
demonstrates stable odometry and consistent map reconstruction across long
bridge structures, which are challenging due to repetitive geometric patterns
and limited features.}
\label{fig:quanlitative_helipr}
\end{figure}

\begin{figure}[!t] %Fig4
\centering
\includegraphics[width=1.0\linewidth]{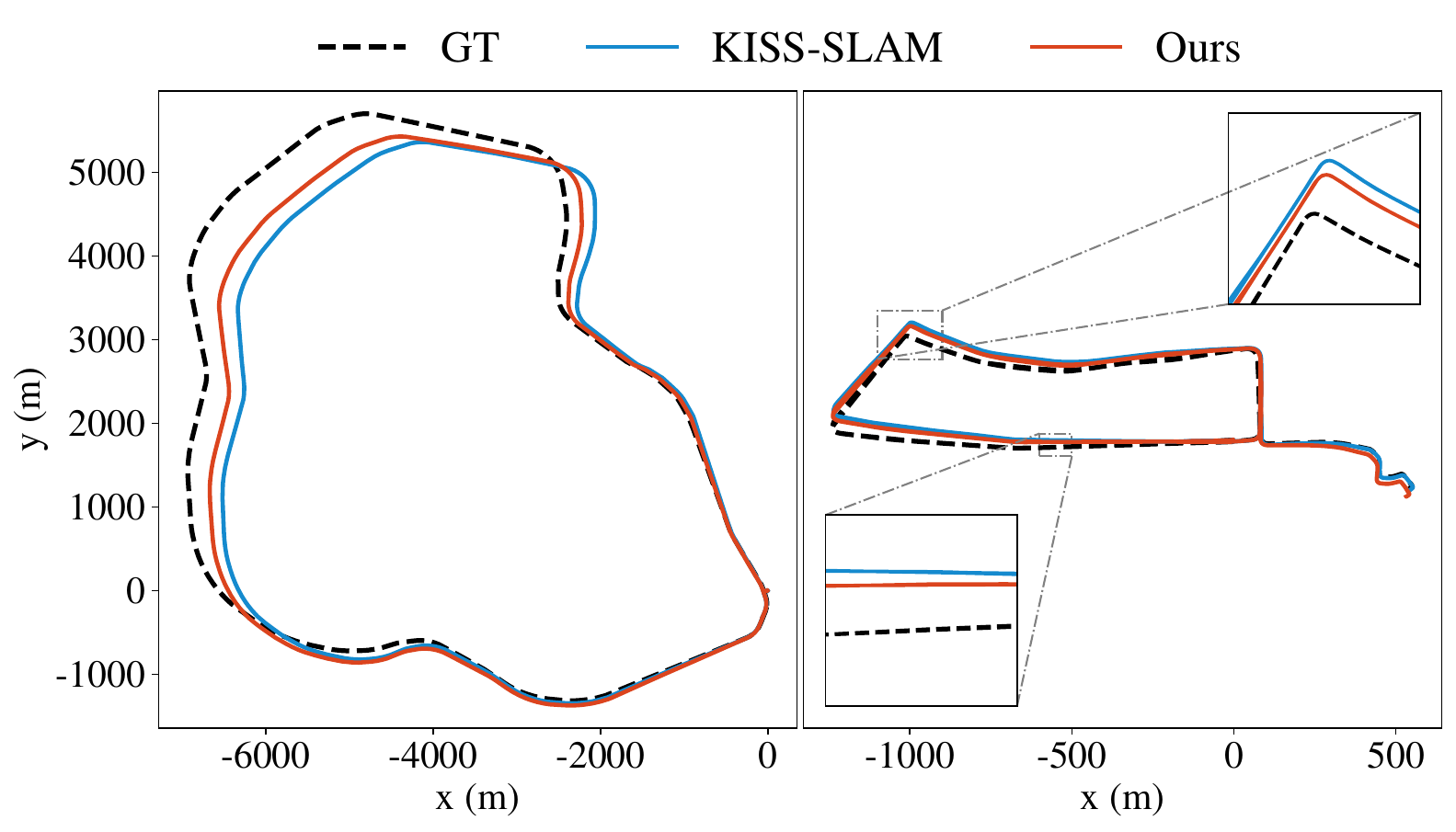}
\caption{Trajectory overlays on Sejong01 (left) and Riverside02 (right). Our
approach aligns more closely with the ground truth (GT), particularly in
challenging curved and loop regions, as shown in the zoomed-in views. 
}
\label{fig:quanlitative_riverside02_trj}
\end{figure}

After alignment, the local map is updated with the registered scan. To maintain global consistency, we integrate a loop-closure module that uses density-preserving BEV projections. Each local map is projected onto the ground-aligned plane, from which ORB descriptors are extracted and stored in a database. Candidate matches are validated through RANSAC-based 2D alignment before being inserted as constraints into the global pose graph. This back-end optimization corrects long-term drift and enforces map consistency across revisits.

% \begin{figure}[h] %Fig3
% \centering
% \includegraphics[width=1\linewidth]{figures/hkiss_slam_bridge04_v5.pdf}
% \caption{Mapping results on the HeLiPR Bridge (Avia04) sequence. Our method
% demonstrates stable odometry and consistent map reconstruction across long
% bridge structures, which are challenging due to repetitive geometric patterns
% and limited features.}
% \label{fig:quanlitative_helipr}
% \end{figure}

% \begin{figure}[h] %Fig4
% \centering
% \includegraphics[width=1.0\linewidth]{figures/sejong_riverside_trajectory_raw_times_tight_2zooms_fix_truefont.pdf}
% \caption{Trajectory overlays on Sejong01 (left) and Riverside02 (right). Our
% approach aligns more closely with the ground truth (GT), particularly in
% challenging curved and loop regions, as shown in the zoomed-in views. 
% }
% \label{fig:quanlitative_riverside02_trj}
% \end{figure}

% \begin{figure}[h]
%     \centering
%     \includegraphics[width=\columnwidth]{figures/Roundabout02_hkiss_slam_v4.pdf}
%     \caption{Qualitative results on the Roundabout02 sequence. KISS-SLAM suffers from drift and map distortions, while our method maintains stable odometry and reconstructs the roundabout with higher consistency.}
%     \label{fig:roundabout02}
% \end{figure}
 
\begin{figure}[h] %Fig5
\centering
% width=0.4\linewidth nghĩa là ảnh chiếm 40% chiều rộng dòng chữ
\includegraphics[width=1\linewidth]{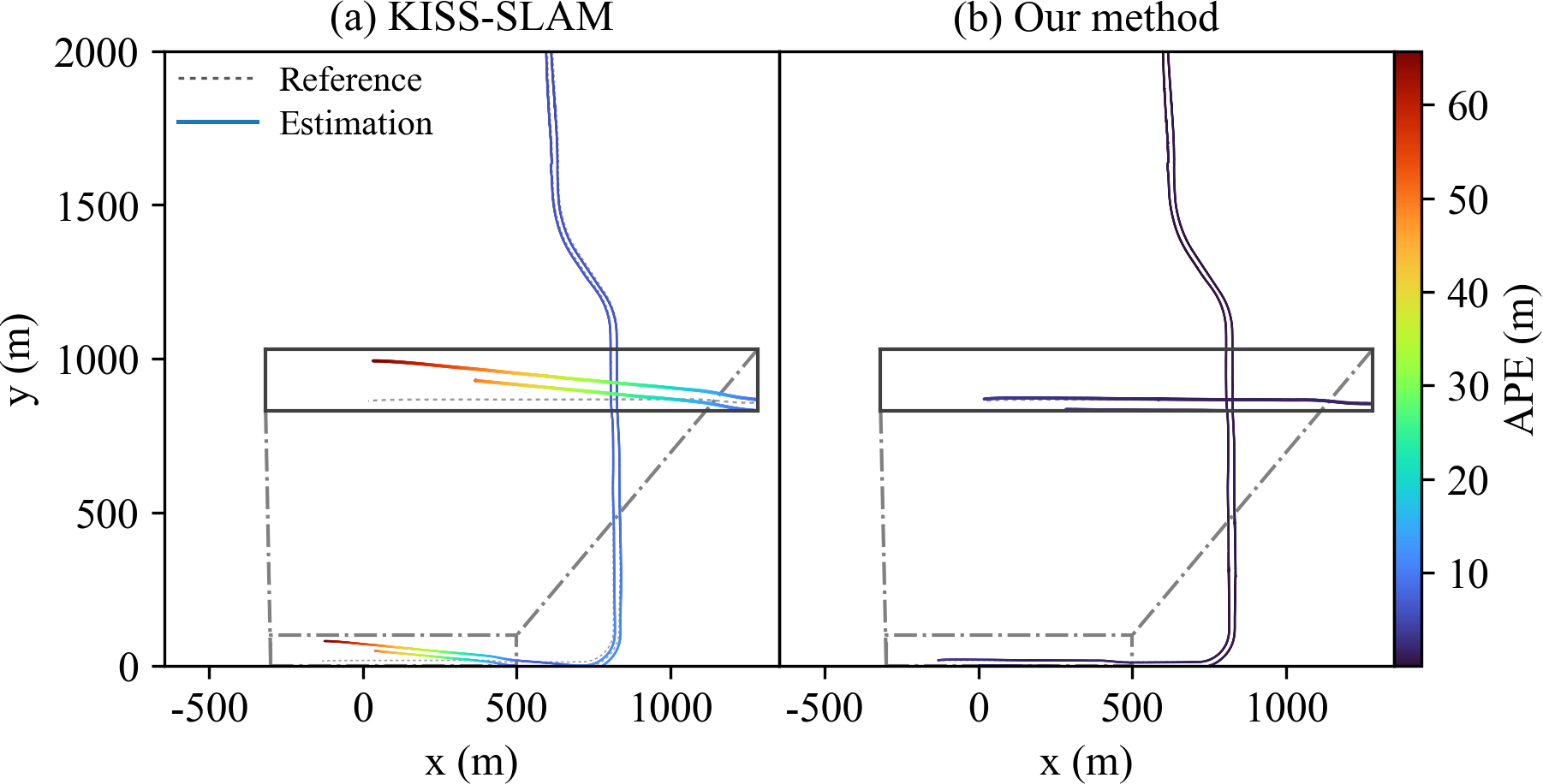}
\caption{An evaluation on the MathildaAVE sequence. Our methodology consistently
yields lower error than the baseline.}
\label{fig:riverside_ape_compare}
\end{figure}

\begin{figure}[h] %Fig6
\centering
\includegraphics[width=1.0\linewidth]{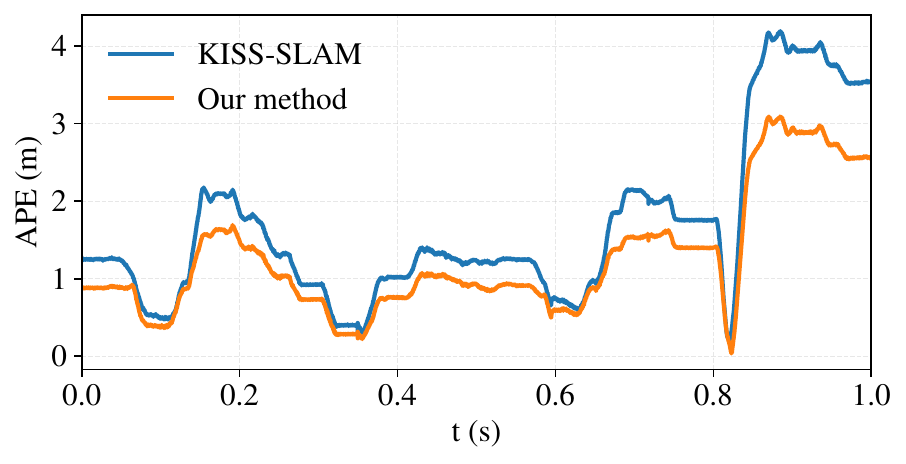}
\caption{The APE of raw trajectories on the SanJose-Downtown dataset.
\paperName~maintains consistently smaller drift in the raw trajectory error,
especially at segments with large error spikes.}
\label{fig:ape_comparison_all}
\end{figure}

\begin{table}[h]
\centering
\resizebox{\columnwidth}{!}{%
\begin{tabular}{lcllcllcllcll}
\hline
\multicolumn{1}{c|}{\multirow{2}{*}{\textbf{Method}}} & \multicolumn{3}{c|}{\textbf{KAIST}} & \multicolumn{3}{c|}{\textbf{DCC}} & \multicolumn{3}{c|}{\textbf{Riverside}} & \multicolumn{3}{c}{\textbf{Sejong}} \\
\multicolumn{1}{c|}{}                                 & \multicolumn{3}{c|}{ATE (m)}        & \multicolumn{3}{c|}{ATE (m)}      & \multicolumn{3}{c|}{ATE (m)}            & \multicolumn{3}{c}{ATE (m)}         \\ \hline
KISS-SLAM~\cite{guadagnino2025kiss}                                             & \multicolumn{3}{c}{2.96}            & \multicolumn{3}{c}{3.68}          & \multicolumn{3}{c}{8.34}                & \multicolumn{3}{c}{\uline{178.34}}    \\
PIN-SLAM~\cite{pan2025pings}                                              & \multicolumn{3}{c}{\textbf{2.42}}   & \multicolumn{3}{c}{\uline{3.57}}    & \multicolumn{3}{c}{\uline{7.55}}          & \multicolumn{3}{c}{782.96}          \\
CT-ICP~\cite{dellenbach2022ct}                                                & \multicolumn{3}{c}{2.76}      & \multicolumn{3}{c}{4.03}          & \multicolumn{3}{c}{8.37}                & \multicolumn{3}{c}{-}               \\
MULLS~\cite{pan2021mulls}                                                 & \multicolumn{3}{c}{34.21}           & \multicolumn{3}{c}{27.30}         & \multicolumn{3}{c}{66.60}               & \multicolumn{3}{c}{1082.55}         \\
\paperName~(ours)                                                  & \multicolumn{3}{c}{\uline{2.71}}      & \multicolumn{3}{c}{\textbf{3.56}} & \multicolumn{3}{c}{\textbf{7.50}}       & \multicolumn{3}{c}{\textbf{146.51}} \\ \hline
\end{tabular}%
}
\caption{A performance comparison of \paperName~and the baselines on the Mulran
dataset. A lower ATE indicates better localization accuracy. The best and second
best performing methods are reported in \textbf{bold} and \uline{underline},
respectively.}
\label{tab:mulran}
\end{table}

\begin{table*}[h]
\centering
\resizebox{\textwidth}{!}{%
% \footnotesize
\begin{tabular}{lcccccccccccc}
\hline
\multicolumn{1}{l|}{\multirow{2}{*}{KITTI-corrected}} & \multicolumn{1}{c|}{00} & \multicolumn{1}{c|}{01} & \multicolumn{1}{c|}{02} & \multicolumn{1}{c|}{03} & \multicolumn{1}{c|}{04} & \multicolumn{1}{c|}{05} & \multicolumn{1}{c|}{06} & \multicolumn{1}{c|}{07} & \multicolumn{1}{c|}{08} & \multicolumn{1}{c|}{09} & \multicolumn{1}{c|}{10} & AVG \\
\multicolumn{1}{l|}{} & \multicolumn{1}{c|}{ATE (m)} & \multicolumn{1}{c|}{ATE (m)} & \multicolumn{1}{c|}{ATE (m)} & \multicolumn{1}{c|}{ATE (m)} & \multicolumn{1}{c|}{ATE (m)} & \multicolumn{1}{c|}{ATE (m)} & \multicolumn{1}{c|}{ATE (m)} & \multicolumn{1}{c|}{ATE (m)} & \multicolumn{1}{c|}{ATE (m)} & \multicolumn{1}{c|}{ATE (m)} & \multicolumn{1}{c|}{ATE (m)} & ATE (m) \\ \hline
KISS-SLAM~\cite{guadagnino2025kiss}& 0.89 & 3.11 & 1.92 & 0.67 & 0.38 & 0.67 & 0.32 & 0.33 & 2.55 & 1.25 & 0.71 & 1.16 \\
GenZ-ICP~\cite{lee2024genz}& 5.19 & 25.47 & 11.19 & 3.27 & 0.90 & 1.50 & 0.86 & 0.67 & 4.46 & 3.07 & 2.32 & 5.36 \\
\paperName~(ours) & \textbf{0.85} & \textbf{2.75} & \textbf{1.89} & \textbf{0.51} & \textbf{0.24} & \textbf{0.56} & \textbf{0.31} & \textbf{0.31} & \textbf{2.40} & \textbf{1.15} & \textbf{0.69} & \textbf{1.06} \\ \hline
\end{tabular}%
}
\caption{Quantitative evaluations on the KITTI-corrected benchmark using the ATE
in meters (m). Compared to KISS-SLAM and GenZ-ICP, \paperName~yields lower
average error and improved performance across the majority of sequences.}
\label{tab:kitti}
\end{table*}

\begin{table*}[h]
\centering
% Tăng khoảng đệm cột để chữ nằm chính giữa ô, tránh bị lệch trái
\setlength{\tabcolsep}{15pt} 
\resizebox{\textwidth}{!}{%
\begin{tabular}{lccccccccc}
\hline
\multicolumn{1}{l|}{\multirow{2}{*}{Method}} & \multicolumn{1}{c|}{Bridge} & \multicolumn{1}{c|}{Bridge} & \multicolumn{1}{c|}{Bridge} & \multicolumn{1}{c|}{Round} & \multicolumn{1}{c|}{Round} & \multicolumn{1}{c|}{Round}  & \multicolumn{1}{c|}{Town} & \multicolumn{1}{c|}{Town} & Town          \\
\multicolumn{1}{l|}{}                        & \multicolumn{1}{c|}{Aeva}   & \multicolumn{1}{c|}{Avia}   & \multicolumn{1}{c|}{Ouster} & \multicolumn{1}{c|}{Aeva}  & \multicolumn{1}{c|}{Avia}  & \multicolumn{1}{c|}{Ouster} & \multicolumn{1}{c|}{Aeva} & \multicolumn{1}{c|}{Avia} & Ouster        \\ \hline
KISS-SLAM~\cite{guadagnino2025kiss}          & \uline{98.61}               & 148.88                      & 19.47                       & \uline{6.06}               & 3.84                       & \uline{1.18}                & \uline{14.44}             & 12.01                     & \uline{1.99}  \\
PIN-SLAM~\cite{pan2025pings}                 & -                           & 365.72                      & \uline{19.23}               & -                          & 7.02                       & 1.47                        & 41.19                     & \uline{11.40}             & 2.55          \\
CT-ICP~\cite{dellenbach2022ct}               & -                           & \uline{41.47}               & 576.62                      & 10.04                      & \uline{3.36}               & 1.81                        & 65.29                     & 63.72                     & -             \\
MULLS~\cite{pan2021mulls}                    & 365.06                      & 321.87                      & 52.65                       & 19.08                      & 16.39                      & 2.65                        & 39.82                     & 14.93                     & 4.45          \\
\paperName~(ours)                            & \textbf{57.37}              & \textbf{26.77}              & \textbf{17.51}              & \textbf{5.77}              & \textbf{3.17}              & \textbf{1.08}               & \textbf{11.65}            & \textbf{10.45}            & \textbf{1.91} \\ \hline
\end{tabular}%
}
\caption{Quantitative results on the Bridge, Roundabout and Town sequences of
the HeLiPR dataset using the ATE in meters (m). The best and second best
performing methods are reported in \textbf{bold} and \uline{underline},
respectively.}
\label{tab:heli}
\end{table*}
\begin{table*}[h]
\centering
% Tăng khoảng cách đệm giữa các cột để bảng tự động rộng ra, giúp chữ không bị phóng to
\setlength{\tabcolsep}{14pt} 
\resizebox{\textwidth}{!}{%
\begin{tabular}{lccccccc}
\hline
\multicolumn{1}{l|}{\multirow{2}{*}{Method}} & \multicolumn{1}{c|}{BTS} & \multicolumn{1}{c|}{CP} & \multicolumn{1}{c|}{H237} & \multicolumn{1}{c|}{MAVE} & \multicolumn{1}{c|}{SB} & \multicolumn{1}{c|}{SJD} & \multicolumn{1}{c}{SJD} \\ 
\multicolumn{1}{l|}{} & \multicolumn{1}{c|}{2018-10-12} & \multicolumn{1}{c|}{2018-10-11} & \multicolumn{1}{c|}{2018-10-12} & \multicolumn{1}{c|}{2018-10-12} & \multicolumn{1}{c|}{2018-10-03} & \multicolumn{1}{c|}{2018-10-11 (1)} & 2018-10-11 (2) \\ \hline
KISS-SLAM~\cite{guadagnino2025kiss} & \uline{3.07} & \uline{0.78} & \uline{25.02} & 12.7 & 2.40 & 1.96 & \uline{0.73} \\
PIN-SLAM~\cite{pan2025pings} & 6.10 & \textbf{0.64} & 97.11 & \uline{7.06}  & \uline{2.23}  & 1.86 & 1.15 \\
% SuMa     & 181.19 / 3.55 & - / - & 366.66 / 9.51 & 79.68 / 0.43 & - / - & - / - & - / - \\
CT-ICP~\cite{dellenbach2022ct}   & 10.81  & 0.97 & 258.87 & 61.39 & 667.28 & \textbf{0.78} & 1.09 \\
MULLS~\cite{pan2021mulls}   & 104.14 & 47.03 & 354.21  & 182.59 & - & 13.38  & 11.41 \\
\paperName~(ours)     & \textbf{2.63} & \textbf{0.64}  & \textbf{24.41} & \textbf{4.01} & \textbf{2.07} & \uline{1.46} & \textbf{0.63 } \\ \hline
\end{tabular}%
}
\caption{Quantitative results on the Apollo dataset using the ATE in meters (m).
The best and second best performing methods are reported in \textbf{bold} and
\uline{underline}, respectively.}
\label{tab:appolo}
\end{table*}
% \begin{table*}[h]
% \centering
% \footnotesize % Cài đặt cỡ chữ nhỏ gọn để đồng bộ
% \begin{tabular*}{\textwidth}{@{\extracolsep{\fill}}lccccccc}
% \hline
% \multicolumn{1}{l|}{\multirow{2}{*}{Method}} & \multicolumn{1}{c|}{BTS} & \multicolumn{1}{c|}{CP} & \multicolumn{1}{c|}{H237} & \multicolumn{1}{c|}{MAVE} & \multicolumn{1}{c|}{SB} & \multicolumn{1}{c|}{SJD} & \multicolumn{1}{c}{SJD} \\ 
% \multicolumn{1}{l|}{} & \multicolumn{1}{c|}{2018-10-12} & \multicolumn{1}{c|}{2018-10-11} & \multicolumn{1}{c|}{2018-10-12} & \multicolumn{1}{c|}{2018-10-12} & \multicolumn{1}{c|}{2018-10-03} & \multicolumn{1}{c|}{2018-10-11 (1)} & 2018-10-11 (2) \\ \hline
% KISS-SLAM~\cite{guadagnino2025kiss} & \uline{3.07} & \uline{0.78} & \uline{25.02} & 12.7 & 2.40 & 1.96 & \uline{0.73} \\
% PIN-SLAM~\cite{pan2025pings} & 6.10 & \textbf{0.64} & 97.11 & \uline{7.06}  & \uline{2.23}  & 1.86 & 1.15 \\
% % SuMa     & 181.19 / 3.55 & - / - & 366.66 / 9.51 & 79.68 / 0.43 & - / - & - / - & - / - \\
% CT-ICP~\cite{dellenbach2022ct}   & 10.81  & 0.97 & 258.87 & 61.39 & 667.28 & \textbf{0.78} & 1.09 \\
% MULLS~\cite{pan2021mulls}   & 104.14 & 47.03 & 354.21  & 182.59 & - & 13.38  & 11.41 \\
% \paperName~(ours)     & \textbf{2.63} & \textbf{0.64}  & \textbf{24.41} & \textbf{4.01} & \textbf{2.07} & \uline{1.46} & \textbf{0.63 } \\ \hline
% \end{tabular*}
% \caption{Quantitative results on the Apollo dataset using the ATE in meters (m).
% The best and second best performing methods are reported in \textbf{bold} and
% \uline{underline}, respectively.}
% \label{tab:appolo}
% \end{table*}
\begin{table*}[th]
\centering
\label{tab:frontend_ablation}
\resizebox{\textwidth}{!}{%
% \footnotesize
\begin{tabular}{lccccccccccc}
\hline
\multicolumn{1}{l|}{\multirow{2}{*}{Method}} & \multicolumn{1}{c|}{0} & \multicolumn{1}{c|}{1} & \multicolumn{1}{c|}{2} & \multicolumn{1}{c|}{3} & \multicolumn{1}{c|}{4} & \multicolumn{1}{c|}{5} & \multicolumn{1}{c|}{6} & \multicolumn{1}{c|}{7} & \multicolumn{1}{c|}{8} & \multicolumn{1}{c|}{9} & 10 \\
\multicolumn{1}{l|}{} & \multicolumn{1}{c|}{ATE (m)} & \multicolumn{1}{c|}{ATE (m)} & \multicolumn{1}{c|}{ATE (m)} & \multicolumn{1}{c|}{ATE (m)} & \multicolumn{1}{c|}{ATE (m)} & \multicolumn{1}{c|}{ATE (m)} & \multicolumn{1}{c|}{ATE (m)} & \multicolumn{1}{c|}{ATE (m)} & \multicolumn{1}{c|}{ATE (m)} & \multicolumn{1}{c|}{ATE (m)} & ATE (m) \\ \hline
GenZ-ICP~\cite{lee2024genz} & 4.72 & 17.21 & 15.41 & 3.55 & \uline{0.64} & 1.62 & 1.16 & 0.73 & 5.62 & 3.64 & 2.16 \\
Point-to-point only & \uline{3.50} & 3.15 & \textbf{6.46} & \uline{0.52} & \textbf{0.31} & 1.49 & 0.42 & \uline{0.31} & \uline{2.20} & 1.31 & \uline{0.71} \\
Point-to-plane only & 3.96 & \textbf{2.49} & \uline{6.63} & 0.60 & 2.41 & \uline{1.45} & \textbf{0.40} & 0.39 & 2.41 & \textbf{0.88} & 0.79 \\
% Fixed-weight hybrid & \textbf{3.41} & \uline{2.58} & 6.66 & \textbf{0.51} & \textbf{0.30} & \uline{1.35} & \uline{0.41} & 0.32 & 2.26 & 1.22 & 0.76 \\
Adaptive hybrid (ours) & \textbf{3.46} & \uline{2.70} & 6.64 & \textbf{0.51} & \textbf{0.31} & \textbf{1.32} & \uline{0.41} & \textbf{0.30} & \textbf{2.19} & \uline{1.14} & \textbf{0.67} \\ \hline
\end{tabular}%
}
\caption{A front-end odometry ablation study on KITTI sequences. A lower ATE
indicates better localization accuracy. The best and second best performing
methods are reported in \textbf{bold} and \uline{underline}, respectively.}
\label{tab:frontend_ablation}
\end{table*}

\begin{table}[h]
\centering
\resizebox{\columnwidth}{!}{%
\begin{tabular}{lccccc}
\hline
\multicolumn{1}{l|}{KITTI-corrected} & \multicolumn{1}{c|}{\begin{tabular}{@{}c@{}}Neighbor \\ Search\end{tabular}} & \multicolumn{1}{c|}{PCA} & \multicolumn{1}{c|}{ICP} & \multicolumn{1}{c|}{Backend} & \multicolumn{1}{c}{\textbf{Total}} \\ \hline
Sequence 00 & 5.92 & 1.69 & 0.91 & 8.72 & 17.24 \\
Sequence 01 & 8.68  & 2.50  & 1.07 & 11.36  & 23.61   \\
Sequence 02 & 5.70 & 1.65 & 0.85 & 9.48 & 17.68 \\
Sequence 03 & 8.37 & 2.4 & 1.06 & 9.91 & 21.74 \\
Sequence 04 & 7.01 & 1.98 & 0.97 & 11.24 & 21.20 \\
Sequence 05 & 5.95 & 1.71 & 0.86 & 9.56 & 18.08 \\
Sequence 06 & 8.18 & 2.35 & 1.03 & 12.04 & 23.60 \\
Sequence 07 & 7.12 & 2.04 & 0.97 & 7.24 & 17.37 \\
Sequence 08 & 7.64 & 2.19 & 1.02 & 10.41 & 21.26 \\
Sequence 09 & 6.97 & 2.03 & 0.96 & 10.23 & 20.19 \\
Sequence 10 & 5.06 & 1.47 & 0.81 & 9.00 & 16.34 \\
\textbf{Average} & 6.96 & 2.00 & 0.96 & 9.93 & 19.85 \\ \hline
\end{tabular}%
}
\caption{A runtime (ms) breakdown of the front-end and back-end modules on the KITTI sequences.
}
\label{tab:runtime}
\end{table}
\label{subsec:runtime_analysis}
\begin{figure}[h] %Fig7
\centering
\includegraphics[width=1.0\linewidth]{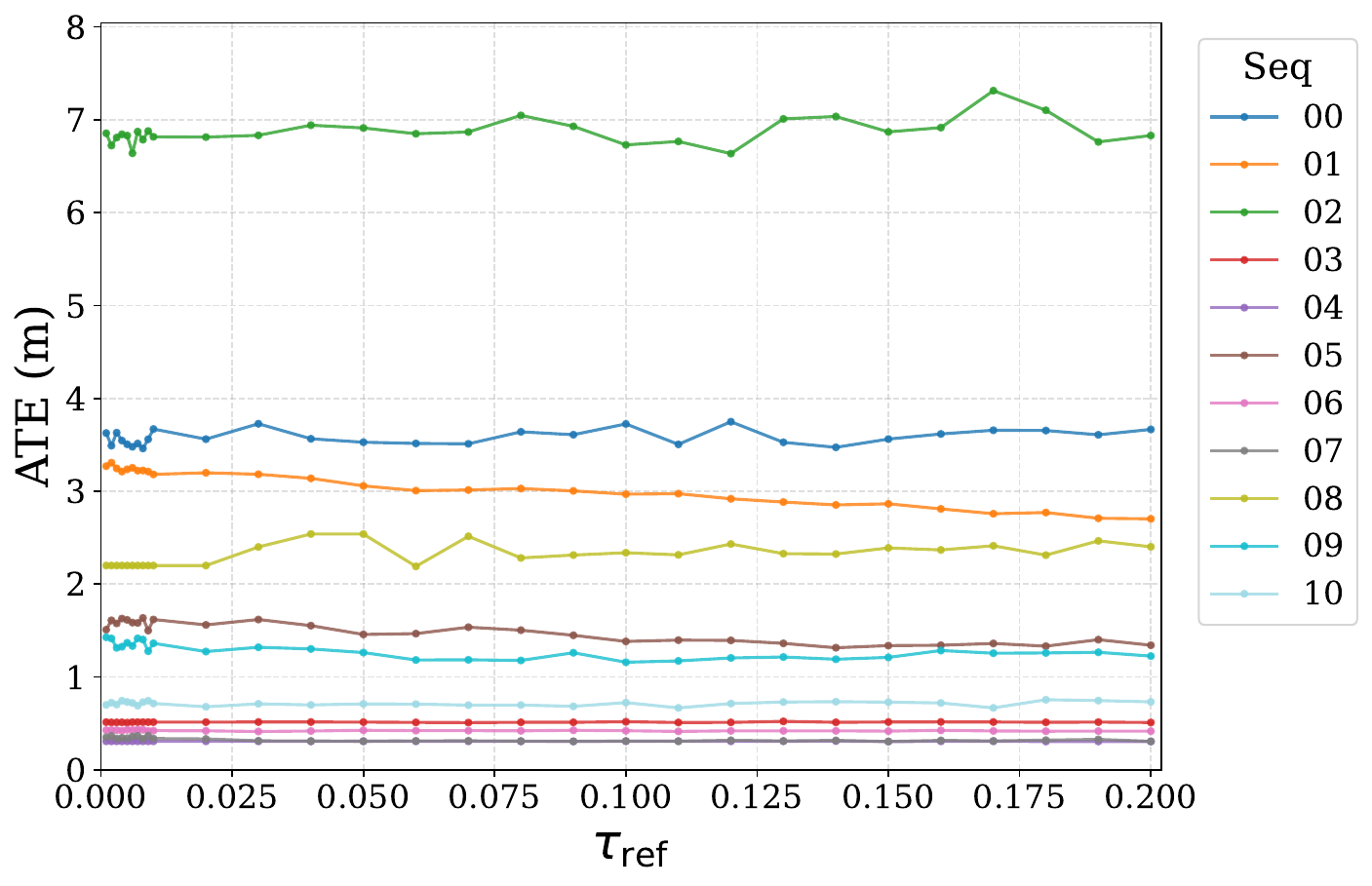}
\caption{The sensitivity of $\tau_{\text{ref}}$ on the KITTI sequences. The ATE
remains stable over a wide parameter range, indicating low sensitivity.
%\cm{Make text in this figure readable (font size should be as big as the captain)}
}
\label{fig:sensitivity_analysis}
\end{figure}

\section{Evaluation}
\label{sec:evalatuion}
In this section, we present an extensive experimental evaluation of the
performance of \paperName. The goal is to examine both odometry accuracy and
mapping consistency under diverse sensing setups and motion conditions. We
validate our methodology on four representative benchmarks:
KITTI~\cite{geiger2013vision}, Apollo~\cite{bakogiannis2019apollo},
MulRan~\cite{kim2020mulran}, and HeLiPR~\cite{jung2024helipr}, across various
LiDAR types including handheld and unmanned aerial vehicle (UAV) scenarios. The
results demonstrate that \paperName~maintains low drift, exhibits strong
adaptability across sensors with minimal or no parameter tuning, and produces
globally consistent maps that can be directly used for robotic navigation and
inspection tasks. 

\subsection{Experimental Setup}
\label{subsec:experimental_setup}
We report odometry accuracy using the absolute trajectory error (ATE), which is
computed after performing a global $SE(3)$ alignment of the estimated trajectory
to the ground truth via the \textit{evo} package~\cite{grupp2017evo}.
Additionally, we evaluate using the KITTI odometry drift metric. Ground-truth
poses are taken from the official releases of each dataset. We start from a
single default configuration and allow limited tuning to improve robustness. In
clearly degenerate cases, we allow at most one generic adjustment to the base
planarity threshold. The adaptive-planarity threshold $\tau(n)$ is clamped to
$[\tau_{\min}, \tau_{\max}]$ and only these bounds may be modified. All
experiments were executed on a machine with an Intel i7 CPU with 32 GB of RAM
and an NVIDIA RTX\,3060 GPU.

\subsection{Qualitative Results} 
\label{subsec:qualitative_results}

We evaluate our approach against the KISS-SLAM baseline across multiple
challenging datasets. On the HeLiPR Bridge (Avia04) sequence, highly regular
patterns and sparse UAV LiDAR data cause severe geometric degeneracy. Since 
KISS-SLAM relies solely on point-to-point ICP, it accumulates noticeable drift
and map misalignment. By contrast, our hybrid point-to-point and point-to-plane
alignment, combined with adaptive-planarity classification, successfully
exploits limited structural cues to preserve tight trajectory alignment and map
consistency, as highlighted in Fig.~\ref{fig:quanlitative_helipr}.

Fig.~\ref{fig:quanlitative_riverside02_trj} demonstrates our method's tighter
ground-truth alignment on the Sejong01 and Riverside02 datasets, successfully
mitigating trajectory deviations during curved loops, sharp turns, and loop
closures. Absolute pose error (APE) evaluations on the MathildaAVE dataset in
Fig.~\ref{fig:riverside_ape_compare} show that our pipeline prevents the gradual
drift typical of KISS-SLAM along extended straight-line segments. Finally,
Fig.~\ref{fig:ape_comparison_all} shows the raw APE profiles from the
SanJose-Downtown sequence and confirms that \paperName~yields a significantly
lower overall error magnitude. Even during sudden error spikes, our trajectory
stabilizes and recovers faster, proving its resilience in environments
characterized by strong rotational motion and repetitive structures.

\subsection{Quantitative Results}
\label{subsec:quantitative_results}

We first evaluate our method and compare our performance against
state-of-the-art pipelines on the MulRan benchmark. As shown in
Table~\ref{tab:mulran}, \paperName~achieves an average ATE competitive with
CT-ICP and PIN-SLAM, while substantially outperforming MULLS and KISS-SLAM,
especially on the high-speed Sejong sequence. Highly dynamic downtown areas and
featureless riverside routes (DCC and Riverside sequences) remain challenging
for purely geometric odometry due to ICP degeneracy. Nevertheless, allowing
minor adjustments to the base planarity threshold is sufficient to stabilize the
optimization in these structure-scarce environments, entirely avoiding the need
for extensive dataset-specific tuning.

We then evaluate our approach on the KITTI dataset to assess global trajectory
consistency via the ATE. As reported in Table~\ref{tab:kitti},
\paperName~consistently yields a lower average ATE than the KISS-SLAM baseline.
Although absolute error reduction is moderate, the proposed hybrid alignment and
adaptive-planarity threshold clearly improve robustness across diverse geometric
conditions without sacrificing accuracy. We also compare against GenZ-ICP, a
front-end-only system that naturally accumulates higher global drift due to the
lack of loop closure. Unlike GenZ-ICP, our pipeline is explicitly designed to
minimize global trajectory errors while preserving the standard ICP efficiency.

Table~\ref{tab:heli} highlights the pipeline's robustness across different
sensors and motion diversity. It consistently ranks in the top tier across
sequences, particularly with Ouster LiDARs. This indicates that it can handle
sparse scans and rapid rotations without sensor-specific overfitting. On the
large-scale Apollo dataset in Table \ref{tab:appolo}, our method maintains high
accuracy under dense traffic, on wide multi-lane roads, and across cross-day
traversals. This stability indicates that the hybrid formulation effectively
mitigates the effects of transient dynamic objects and adapts well to appearance
changes. Notably, we achieve these results by limiting the base planarity
threshold during degenerate sequences, confirming our system's practicality for
real-world urban deployments. 

\subsection{Ablation Study and Sensitivity Analysis}
\label{subsec:ablation_study_and_sensitivity_analysis}
Table~\ref{tab:frontend_ablation} presents the front-end ablation study on KITTI
sequences 00–10. While single geometric constraints are highly
environment-dependent (point-to-plane excels in sequences 01, 06, and 09, and
point-to-point in sequences 02 and 04), our adaptive hybrid method exhibits
superior robustness. It achieves the best and second-best ATE in 7 and 3,
respectively, out of 11 sequences. The method also notably reduces sequence 03's
ATE to 0.51 m, compared with 3.55 m for the GenZ-ICP baseline. These results
confirm that dynamically weighting loss functions based on local geometry
effectively mitigates drift across complex trajectories.

We observe in Fig.~\ref{fig:sensitivity_analysis} that varying
$\tau_{\mathrm{ref}}$ over a broad interval results in only minor and smooth
changes in ATE. No sharp performance collapse is observed, even away from the
optimal setting. This suggests that the adaptive formulation mitigates the
strong parameter sensitivity typically associated with fixed planarity
thresholds and maintains stable behavior across different environments.

\subsection{Runtime Analysis}
We assess the runtime performance of \paperName~on the KITTI sequence. Our
method achieves an average runtime of 19 ms per scan, compared to 13 ms for
KISS-SLAM. However, the system still operates above 50 Hz, thereby enabling
real-time applications. The runtime breakdown indicates that the proposed
front-end is highly computationally efficient, as shown in
Table~\ref{tab:runtime}. The per-frame processing time remains below the scan
period of typical LiDAR sensors, confirming that integrating adaptive-planarity
estimation and hybrid alignment preserves real-time performance.

% \begin{table}[h!]
%     \centering
%     \caption{Front-end Odometry Ablation Study on Kitti
% sequence 00} \cm{Please explain and highlight the performance of different approaches in the ablation study similarly to other tables}
%     \begin{tabular}{l c}
%         \toprule
%         \textbf{Method} & \textbf{ATE} \\
%         \midrule
%         Point-to-point (KISS-ICP)  &  4.924   \\
%         Point-to-plane only        &  4.646   \\
%         Fixed-weight hybrid        & 4.773 \\
%         \textbf{Adaptive hybrid (ours)} & 4.424    \\
%         \textbf{GenZ-ICP}          &  5.241   \\
%         \bottomrule
%     \end{tabular}
% \end{table}

% \begin{table}[t]
% \centering
% \begin{minipage}{0.48\linewidth}
% \centering
% \caption{Table 1}
% \begin{tabular}{c c}
%     % \toprule
%     \textbf{Method} & \textbf{ATE / Drift} \\
%     % \midrule
%     Point-to-point (KISS-ICP)  &  \\
%     Point-to-plane only        &  \\
%     Fixed-weight hybrid        &  \\
%     \textbf{Adaptive hybrid (ours)} &  \\
%     \textbf{GenZ-ICP}          &  \\
%     % \bottomrule
% \end{tabular}
% \end{minipage}
% \hfill
% \begin{minipage}{0.48\linewidth}
% \centering
% \caption{Table 2}
% \begin{tabular}{c c}
%         % \toprule
%         \textbf{Module} & \textbf{Time (ms)} \\
%         % \midrule
%         Neighbor search  & 5.92 \\
%         PCA + planarity  & 1.69 \\
%         ICP optimization & 0.91 \\
%         % \midrule
%         \textbf{Total}   & 8.52 \\
%         % \bottomrule
% \end{tabular}
% \end{minipage}
% \end{table}
\section{Conclusion}
\label{sec:conclusion}
In this paper, we presented \paperName, a LiDAR SLAM framework that provides
both a minimalist philosophy for real-time applications and a novel adaptive
hybrid ICP. Our approach aims to dynamically balance point-to-point and
point-to-plane via a neighborhood-size adaptive planarity threshold, achieving stable alignment in
both structured and degenerate environments without relying on complex feature engineering. Integrated with loop closure and pose graph optimization, this front-end enables globally consistent mapping over long trajectories. This research demonstrates that \paperName~outperforms or matches
strong baselines across widely used benchmarks. Moreover, \paperName~achieves
notable robustness in degenerate scenarios, allows for cross-sensor
generalization without requiring dataset-specific tuning, and provides real-time
efficiency on commodity hardware. Future work will explore pairing this adaptive
hybrid ICP with complementary sensing modalities and test operation in long-term
autonomy scenarios. 

\bibliographystyle{IEEEtran}
\bibliography{IEEEabrv, 09_references}

\end{document}